# The Nonverbal Gap: Toward Affective Computer Vision for Safer and More Equitable Online Dating

Ratna Kandala[1]
University of Kansas
ratnanirupama@gmail.com

Niva Manchanda[2]
University of Kansas
nmanchanda@ku.edu

Akshata Kishore Moharir[3]
University of Maryland
akshatankishore5@gmail.com

## Abstract

*Online dating has become the dominant way romantic relationships begin, yet current platforms strip the nonverbal cues: gaze, facial expression, body posture, response timing, that humans rely on to signal comfort, disinterest, and consent, creating a communication gap with disproportionate safety consequences for women. We argue that this gap represents both a technical opportunity and a moral responsibility for the computer vision community, which has developed the affective tools, facial action unit detection, gaze estimation, engagement modeling, and multimodal affect recognition, needed to begin addressing it, yet has largely ignored the dating domain as a research context. We propose a fairness-first research agenda organized around four capability areas: real-time discomfort detection, engagement asymmetry modeling between partners, consent-aware interaction design, and longitudinal interaction summarization, each grounded in established CV methodology and motivated by the social psychology of romantic communication. We argue that responsible pursuit of this agenda requires purpose-built datasets collected under dyadic consent protocols, fairness evaluation disaggregated across race, gender identity, neurotype, and cultural background, and architectural commitments to on-device processing that prevent affective data from becoming platform surveillance infrastructure. This vision paper calls on the WICV community, whose members are uniquely positioned to understand both the technical opportunity and the human stakes, to establish online dating safety as a first-class research domain before commercial deployment outpaces ethical deliberation.*

## 1. Introduction

Online dating has become the dominant way people meet romantic partners across the US and Europe [21, 39, 90], yet the communication interfaces supporting it are remarkably primitive from a perceptual standpoint. Face-to-face romantic interaction is richly multimodal: people read microexpressions, track gaze aversion, register hesitation, and interpret body posture to continuously calibrate comfort, interest, and distress [27, 29, 32, 82]. Text-based messaging and emoji reduce this rich signal to a narrow, low-bandwidth channel [8, 58, 88], while even video dating platforms process their streams for compression and presentation rather than affective content. The result is a profound modality mismatch between how humans are equipped to communicate romantically and what current platforms actually support.

This gap is not perceptually neutral-it is socially consequential. Research in social psychology consistently shows that women disproportionately manage unwanted social advances in dating contexts and rely heavily on nonverbal channels to signal disinterest safely and without confrontation [60, 82]. When those channels are absent, discomfort signals go unread, unwanted escalation goes unchecked, and the burden falls on the more vulnerable party to communicate refusal explicitly in a medium that was never designed for it (e.g., matrimonial sites[67, 81]). The absence of an affective signal does not affect all users equally-it amplifies existing vulnerabilities.

Meanwhile, the computer vision community has made substantial advances in precisely the technologies that could address this gap. Facial action unit detection [3], gaze estimation [97], emotion recognition [59], and multimodal affective modeling [64] have each matured considerably over the past decade. Yet almost none of this work has been examined through the lens of dating communication, consent signaling, or intimate interaction safety. There is no established research agenda connecting affective CV to online dating as a sociotechnical system, and no fairness literature examining how such tools would perform across gender, age, race, and sexual orientation in this specific context.

This paper argues that this gap is both a technical opportunity and a moral responsibility for the CV community. We ask: what nonverbal signals are most diagnostic of comfort, disinterest, and distress in romantic communication contexts, and can they be reliably detected from video?

How does the absence of these signals in platform-mediated dating disproportionately affect women's safety and autonomy? And what would a fairness-first, consent-aware affective vision research agenda for online dating actually look like? We outline this agenda, identify the key technical and ethical challenges, and call on the CV community to take ownership of it.

## 2. The Nonverbal Signal Gap in Dating Communication

Human communication in romantic contexts, such as face-to-face dating, is fundamentally nonverbal. Long before words are exchanged, people rely on a continuous stream of physical signals such as direction of gaze, postural orientation, proximity, touch initiation, and the timing and quality of facial expressions - to assess attraction, signal availability, and negotiate comfort [9, 27, 32]. These signals are not supplementary to romantic communication; they are its primary substrate [50, 75]. They are often subtle, rapid, and partly involuntary, and play a central role in how people convey both romantic interest and rejection. [29] demonstrated that non-verbal cues during courtship interactions predicted romantic interest more reliably than verbal content alone, while [70] showed that sustained gaze, open body orientation, space maximization were systematically deployed by men seeking to attract female attention. Flirting itself is distinguished from ordinary friendly interaction almost entirely through nonverbal modulation of otherwise neutral verbal content [31].

Of particular interest is the role of non-verbal cues in signaling disinterest and discomfort. Literature from social psychology has long established that explicit verbal refusal is not the primary mechanism through which disinterest is communicated, particularly for women navigating unwanted advances [60, 82]. Gaze aversion, reduced smiling, postural withdrawal, and response latency elongation are among the signals people, disproportionately women, deploy to communicate disengagement without direct confrontation [53]. Abbey's [1] foundational work demonstrated that men systematically misattribute friendly behavior as sexual interest, a misperception that nonverbal cues in face-to-face contexts can partially correct but that text-based communication eliminates entirely. The consequences of this signal loss are not trivial: [22] linked the misperception of sexual intent directly to coercive behavior.

App-based communication strips most of this away, leaving primarily text, emojis, and occasional static images. Work by Walther [91, 92] established that computer-mediated communication, even when asynchronous text is replaced by richer modalities, fundamentally alters the interpersonal dynamics of interaction. Emoji and emoticons, the primary affective tools available for dating app users, are a poor substitute for genuine nonverbal signals. While they can convey broad valence, they require active, explicit, deliberate input: they cannot capture the involuntary microexpressions, gaze behavior, or postural shifts that carry the most diagnostically useful information [16, 73]. Kruger et al. [52] demonstrated that communicators dramatically overestimate how well their emotional tone is conveyed in text, suggesting that dating app users are likely operating under a systematic illusion of communicative richness that the channel does not actually support.

Video dating, which expanded dramatically during covid-19 pandemic and has since become a standard feature, might appear to close this gap. It does not. Even when an audio or video is present, current platforms rarely do anything computationally with the rich nonverbal stream. Video-mediated communication introduces its own set of perceptual distortions: eye contact is structurally impossible due to camera-screen offset [6, 30], spatial faithfulness is compromised by fixed camera angles [61], and the self-view interface creates a self-monitoring burden that disrupts natural expressions. Crucially, no current dating platform processes its video stream for affective content in any user-beneficial way. The result is a structural nonverbal signal gap: key cues about comfort, consent, and interest are missing from the interaction channel. This loss is not psychologically symmetric. Individuals who are less assertive or more conflict-avoidant patterns that are shaped by, but not reducible to, gender socialization often rely on gradual, nonverbal signals to indicate disinterest because saying "no" directly feels risky or costly [43, 44, 87]. For many women and marginalized groups, nonverbal cues provide a safer, lower-conflict way to communicate boundaries in flirtatious or ambiguous situations [41, 85]. When platforms remove or minimize these channels, they disproportionately disadvantage users who depend on them to manage unwanted advances and avoid confrontation. From a computational perspective, this raises concrete questions about which nonverbal signals, if used carefully and with consent, could help recover some of the lost safety and clarity.

This section establishes two claims that motivate the rest of the paper. First, nonverbal cues are not peripheral to romantic communication but central to it, particularly for the signaling of discomfort and disinterest, and their absence creates measurable communicative and safety harms. Second, video dating platforms are technically capable of capturing these cues but currently do not, creating a tractable research gap that the computer vision community is uniquely positioned to address.

## 3. What Affective Vision Could Offer: A Proposed Research Agenda

Having established that nonverbal signal is both essential to romantic communication and largely absent from current platforms, we now outline a concrete research agenda for affective computer vision in this domain. While this paper foregrounds visual modalities consistent with its CV framing, we acknowledge that audio prosody — pitch variation, speech rate, pause duration, hesitation markers — and turn-taking dynamics carry strong complementary signal for discomfort and engagement [54, 79]. A complete system should be multimodal by design [65]. We treat the visual channel as the primary contribution of this agenda while explicitly flagging audio and linguistic features as natural and necessary extensions.

We organize this agenda around four interconnected capability areas: discomfort and disinterest detection, engagement asymmetry modeling, consent-aware interaction design, and longitudinal interaction summarization. For each, we describe the technical approach, identify the key open research questions, and discuss what user-facing applications would look like in practice. Throughout, we emphasize that the goal of this agenda is user empowerment, giving individuals more legible information about their own interactions, rather than platform surveillance or autonomous decision-making.

### 3.1. Discomfort and Disinterest Detection

The most direct application of affective CV to dating safety is the real-time detection of negative affect and disengagement during video interactions. Facial Action Coding Systems (FACS), formalized by [19] and now automatable at high accuracy through tools such as OpenFace 2.0 [3], provides a theoretically grounded and expression-agnostic representation of facial behavior. Specific AU patterns such as AU4(brow lowering), AU15 (lip corner depression), AU17 (chin raising) are reliably associated with discomfort in neurotypical Western populations [19, 86]. Gaze detection, detectable through standard webcams, is among the most consistent non-verbal markers of social disengagement and discomfort [46, 97]. While no prototype is presented here, individual components have been demonstrated in adjacent domains: OpenFace 2.0 processes facial behavior in real time at 30fps on consumer hardware [3], monocular gaze estimation achieves under 4° mean angular error in unconstrained settings [46], and engagement detection from video has been validated in educational dyadic settings [93]. The open research challenge is therefore not component feasibility but integration, ecological validity in dating contexts, and equitable performance across demographic groups.

We define *discomfort* as a sustained co-occurrence of the above AU patterns persisting beyond two seconds, combined with gaze aversion exceeding the individual baseline by more than 1.5 standard deviations, with labels validated against participant self-report using intraclass correlation coefficients [12, 26]. *Disinterest* is operationalized as reduced response contingency between partners, measured as cross-correlation lag in expressive streams falling below an empirically established threshold [15], combined with reduced head nod frequency relative to individual baseline. *Consent signaling* is treated as a composite behavioral state rather than a single cue [22], requiring dedicated annotation protocols developed in collaboration with social psychologists before any dataset collection begins. We acknowledge that Barrett et al. [5] have raised fundamental challenges to inferring discrete emotional states from facial movements alone, and we therefore treat these AU patterns not as direct emotion readouts but as behavioral correlates requiring contextual validation in the specific dyadic romantic setting. We also note that AU-to-affect mappings carry well-documented cross-cultural validity limitations [45, 57], a challenge we treat as a core fairness requirement in Section 4.4 rather than an implementation detail, and one that any deployed system must address through culturally stratified validation rather than assuming universality.

The key research questions here are not primarily whether individual cues can be detected, but whether they can be reliably interpreted in a naturalistic, unconstrained, and emotionally complex context of dating. Existing affective datasets, such as RECOLA [72] and SEWA DB [51] capture dyadic remote interactions but were not collected in romantic contexts and also do not include consent-relevant behavioral annotations. New data collection, with appropriate ethical oversight and dyadic consent protocols, is needed. Detection also needs to distinguish genuine discomfort from conversational norms such as looking away while thinking, laughing with eyes closed, or cultural variation in baseline gaze behavior [63]- a significant but tractable signal processing challenge.

In terms of application, we envision discomfort detection not as an autonomous intervention but as a user-controlled awareness tool, analogous to heart rate monitor that a user can act on. A simple, private, end-of-interaction summary ("there were several moments where your expressions suggested discomfort, here is where they occurred") could provide the kind of reflective affordance that current platforms entirely lack. Critically, this data would be processed and displayed only to the user whose face is being analyzed, never shared with the other party or the platform.

### 3.2. Engagement and Asymmetry Modeling

Beyond individual affect detection, a particularly promising direction is the modeling of engagement asymmetry between partners, detecting when one party is significantly more invested, attentive, or emotionally activated than the other. Interpersonal synchrony research has established that mutual engagement in dyadic interaction produces measurable behavioral alignment: matched facial expressions, coordinated head movements, synchronized response timing [15, 55],

Conversely, asymmetric engagement, where one party is highly activated and the other is withdrawn, produces a characteristic divergence in these behavioral streams. Operationalizing engagement asymmetry requires empirical decisions on temporal window, synchrony research suggests 30–60 second windows balance sensitivity and noise [15], asymmetry threshold, and individual baseline calibration to avoid penalizing naturally low-expressivity users. We propose that these parameters be established empirically through correlation with self-reported comfort ratings rather than set a priori, and treat them as primary empirical contributions of the benchmark proposed in Section 3.5.

Modeling this asymmetry requires moving beyond single-face analysis to genuine dyadic modeling, tracking both participants simultaneously and computing relational features across the pair rather than treating each face independently. Techniques from group behavior analysis [76] and virtual rapport modeling [40] provide relevant methodological foundations, though they were not developed for romantic dyads and have not been validated in this context.

The research questions here include: Can engagement asymmetry be detected reliably from naturalistic video (speed) dating settings? What level of asymmetry is diagnostically meaningful versus normal variation? What temporal window is appropriate: asymmetry meaningful over seconds, minutes, or the full interaction? How does the model handle turn-taking dynamics where asymmetric attention is structurally expected? And critically, what level of asymmetry is diagnostically meaningful versus normal variation? Answering these questions requires both datasets and close collaboration with social psychologists who study romantic relationships, attraction, and interest signaling in dyadic contexts [68].

### 3.3. Consent-Aware Interaction Design

A third direction moves from detection toward design, using affective CV insights to build platform affordances that make it easier to disengage from uncomfortable interactions without confrontation. This is partly a CV problem and partly an HCI problem, and it requires both communities working together. The core insight is that platforms currently offer only blunt exit mechanisms: unmatching, blocking, and ending a call, which require explicit, socially costly action. There is no soft affordance for signaling reduced interest that does not require the user to initiate a potentially confrontational act.

CV-powered tools could support subtler mechanisms. A detected pattern of sustained disengagement could quietly surface an easy exit option: "would you like to wrap up this call?" without requiring the user to explain or justify. Work on persuasive technology [23] and behavior change design [13] provides a framework for thinking about how such nudges can be designed to feel supportive rather than intrusive. Rader's work [66] on algorithmic transparency is relevant to how any affective processing should be disclosed to users.

The key research questions here are primarily in the HCI domain but depend on CV inputs. What interface designs for surfacing affective feedback are experienced as helpful versus invasive? How should the system communicate uncertainty, what happens when the model is unsure whether discomfort is genuine? How do users from different demographic groups respond to affective nudges, and are there differential effects by gender, age, or cultural background? These questions require user studies with diverse participants, ideally in collaboration with dating platform partners who can support ecologically valid deployment.

Participatory safety research in social matching systems has documented women's concrete design priorities for safer dating interactions [2], and consent technology design with LGBTQ+ stakeholders has identified specific affordances supporting autonomous refusal [99]. The capability areas we propose should be developed in direct dialogue with this literature rather than in isolation."

### 3.4. Longitudinal Interaction Summarization

A fourth and more speculative direction is the post-hoc summarization of interaction dynamics across a video date: giving users a visual and affective "replay" of how the interaction evolved, analogous to how fitness apps provide post-workout analysis. Rather than real-time intervention, this approach surfaces patterns across the full interaction: moments of high mutual engagement, points where one party withdrew, and the temporal trajectory of expressed affect on both sides.

This framing draws on precedents in affective computing for educational and therapeutic contexts [28, 49], where longitudinal behavioral feedback has been shown to support self-awareness and behavior change. In a dating context, the value proposition is less about changing behavior in the moment and more about helping users, particularly those who find it difficult to trust their own read of an interaction, develop a more grounded understanding of what actually happened. For users who have experienced gaslighting or coercive control in past relationships, objective behavioral data about interaction dynamics could be meaningfully supportive. The research questions here include: what visualizations of affective interaction data are interpretable and useful to non-expert users? How should uncertainty in model outputs be communicated so that users do not over-interpret noisy signals? And what are the risks of misuse? Could a bad actor use interaction summaries to refine manipulative techniques? This last question connects directly to Section 5 on privacy and dual-use risk.

### 3.5. Towards a Shared Benchmark

Underpinning all four of these capability areas is a shared need: a purpose-built dataset for affective CV in romantic dyadic interaction. Existing datasets [51, 68, 72] provide partial coverage but were not designed for this purpose, lack consent-relevant annotations, and do not reflect the demographic diversity needed for fair model development. Also, existing dyadic annotation tools, including ELAN [94] and py-feat [10], provide methodological infrastructure that could significantly accelerate benchmark creation without requiring bespoke annotation pipelines.

We call for the creation of a new benchmark dataset collected under rigorous dyadic consent protocols, annotated for discomfort, disinterest, engagement, and consent signaling, and stratified across gender, age, race, sexual orientation, and cultural background. The ethical framework for this collection is itself a research contribution and is discussed further in Section 5.

## 4. Fairness and Equity Considerations

The research agenda outlined in Section 3 is technically promising, but its promise is contingent on a condition that the affective computing community has historically failed to meet: that the systems built perform equitably across the full demographic range of the people who will use them. This section argues that fairness auditing is not an optional addendum to the proposed agenda but a prerequisite for any responsible deployment, and that the specific context of intimate interaction makes the consequences of biased systems particularly severe.

### 4.1. Racial and Skin Tone Bias in Affective CV

The foundational work by Buolamwini and Gebru [7] demonstrated that commercial face analysis systems perform significantly worse on darker-skinned faces and on women, with the worst performance concentrated at their intersection. Subsequent work has confirmed and extended these findings: [71] found that emotion classifiers systematically rated Black faces as expressing more negative emotion than White faces displaying equivalent expressions. [96] documented similar racial performance gaps across multiple commercial affective computing approaches.

In a dating context, these biases are not merely technical imperfections; they translate directly into harmful user experiences. A system that systematically misreads Black women's expressions as more negative would generate false discomfort signals, potentially causing users to distrust interactions that were in fact positive, or causing platforms, if they had access to this data, to down-rank certain users. Even in a fully user-facing, privacy-preserving deployment, a biased system that consistently mischaracterizes a user's own emotional state erodes trust and utility in ways that fall disproportionately on already marginalized users. Any affective CV system deployed in a dating context must therefore demonstrate equitable performance across racial and skin tone groups before deployment, using evaluation protocols that go beyond aggregate accuracy to examine performance at demographic intersections [18, 35].

Age represents a further underexamined bias axis: FER systems perform significantly worse on older adults [48], while major training datasets including AffectNet and FER-2013 lack adequate elderly representation [25]. Given that online dating is increasingly used across wide age ranges, this bias requires explicit attention in dataset collection and evaluation protocols.

### 4.2. Gender and Limits of Binary Classification

Most affective CV systems, including those for emotion recognition, gaze analysis, and engagement modeling, encode gender as a binary covariate used either as a feature or as a stratification variable in evaluation. This becomes problematic for all users but is acutely harmful for trans and nonbinary users. Os Keyes[47] documented how automatic gender recognition systems systematically misclassify trans individuals, producing errors that are not random but systematically correlated with the direction of gender transition. Scheuerman et al.[78] extended this to commercial facial analysis and found consistent, significant misclassification rates for trans and nonbinary faces. Hamidi et al.[33] argued that automatic gender recognition for these populations is not merely inaccurate but structurally reductive- it enforces a binary that many users do not inhabit.

For affective CV in dating contexts this matters in two ways. First, gender is often used as a conditioning variable in affect models, if the underlying gender classifier is unreliable for a significant user population, affect predictions conditioned on it will be systematically distorted. Second, the dating context is one where gender identity is particularly salient and where misclassification by a platform system, even one operating invisibly, could cause real psychological harm. We call for affective dating CV systems to be designed without binary gender assumptions wherever possible, and for evaluation to explicitly include trans and nonbinary participants as a required demographic stratum rather than an afterthought.

### 4.3. Neurodivergent Users

Affective CV systems are trained overwhelmingly on neurotypical expressive behavior, calibrated to the norms of how neurotypical people display and recognize emotion. Autistic individuals, for example, may display flat or atypical facial affect that does not conform to these norms even when experiencing strong emotion [36], while individuals with alexithymia, difficulty identifying and describing emotional states [89], may produce expressions that are gen-uinely disconnected from their internal states in ways that confound model assumptions. A discomfort detection system that flags an autistic user's baseline neutral expression as distress, or that fails to detect genuine discomfort because it is expressed atypically, would be both inaccurate and potentially harmful.

This is not a minor edge case: autism spectrum conditions affect an estimated 1–2% of the population, and neurodivergent individuals disproportionately report difficulty with the social navigation demands of dating [36]. These are users who might benefit most from affective support tool, and who are most at risk of being poorly served by systems trained without them. We call for neurodivergent participants to be explicitly included in dataset collection and for model evaluation to report performance disaggregated by neurotype where ground truth labels can be established.

### 4.4. Cultural Variation in Expression Norms

This concern directly qualifies the AU-based discomfort detection proposed in Section 3.1, where we treat AU patterns as behavioral correlates requiring cross-cultural validation rather than universal emotion readouts [5]. The social norms governing when and how emotion is expressed vary significantly across cultures [20, 57]. Jack [45] demonstrated that the facial expressions of emotion are not fully culturally universal, challenging the Ekman model that underlies most affective CV systems. A gaze aversion that signals discomfort in one cultural context may be a routine politeness norm in another; a smile that indicates genuine warmth in one context may be a social mask in another. Models trained on WEIRD [38] populations will systematically misread users from other cultural backgrounds.

Online dating is a global phenomenon, and even within single platforms, users span enormous cultural diversity. Affective CV systems deployed in this context without cross-cultural validation will produce systematically biased outputs for large portions of their user base. Hence, we also call for dataset collection to be explicitly cross-cultural and for evaluation to report performance disaggregated by cultural background, while recognizing this is a harder problem than demographic stratification within a single cultural context.

### 4.5. Fairness Criteria

Standard fairness metrics from ML literature: demographic parity, equalized odds, calibration [11, 35], provide a starting point but are insufficient in the dating domain without adaptation. False positives and false negatives have asymmetric consequences that differ by user group: a false positive discomfort signal for a user who is comfortable wastes attention and erodes trust; a false negative discomfort signal for an uncomfortable user fails them at a moment of potential vulnerability. These asymmetric harms suggest that standard accuracy-based fairness criteria need to be supplemented with harm-weighted evaluation frameworks that account for the specific social consequences of different error types for different demographic groups. We propose that the field develop such a framework as a priority research contribution, ideally collaborating with social scientists who can quantify the real-world consequences of different error profiles [4]

## 5. Privacy, Consent, and the Limits of the Proposal

Processing live video streams for emotional content represents a significant privacy intervention that demands rigorous ethical safeguards, granular user consent, and architectural protections against misuse. Facial expressions, gaze patterns, and body language reveal intimate psychological states that users may not wish to expose, even to themselves, in real time [14, 62]. This section makes explicit the ethical boundaries of the proposed agenda, addresses its most serious risks, and identifies the open research and regulatory questions that must be resolved before responsible deployment is possible.

### 5.1. User-Facing Tools

This proposal explicitly advocates only for user-facing tools, where affective analysis occurs locally on the user's device and remains visible solely to that individual (e.g., "Your facial tension suggests discomfort, want to pause or end this call?"). Platform-facing applications [69], where emotional data is collected centrally for matching, moderation, monetization, or recommendation, are explicitly excluded due to their high risk of surveillance and exploitation [98].

This distinction has a direct technical corollary: affective processing should occur on-device wherever architecturally feasible. Federated learning [34], on-device inference pipelines, and zero-knowledge proof frameworks [74] offer concrete pathways to ensure that raw emotional data does not leave the user's device.

### 5.2. The Dyadic Consent Problem

Consent in intimate contexts such as dating is particularly complex. Power imbalances, social expectations, and opaque data practices can undermine the conditions necessary for truly voluntary agreement. Informed consent for video processing becomes even more challenging in dyadic interactions [56]. For example, in a video date, both individuals are simultaneously visible, meaning that their facial expressions, gaze patterns, and other behavioral cues may be subject to analysis. This creates a structural asymmetry in consent: one party may agree to have their own video stream processed, while the other may object to the analysis of theirs. Such asymmetries complicate traditional models of individual consent, as the data of multiple individuals are inherently intertwined within a single interaction.

We propose three principles for dyadic consent in this context. First, by default affective processing should be limited strictly to the consenting user's own behavioral stream, with the other party's face either excluded from analysis or processed only for features that do not leave the device. Second, any analysis that incorporates the other party's behavior, such as engagement asymmetry modeling should require that party's separate, explicit, informed consent before activation. Third, consent must be genuinely granular and revocable in real time: users should be able to opt into specific capabilities independently and withdraw at any point without friction or penalty [77].

Dating platforms have faced criticism for treating vague "OK" pop-up acknowledgements as consent for AI training on chat histories and photos [56], demonstrating that the current industry norm for consent is inadequate for high-sensitivity data, highlighting the need for just-in-time, revocable permissions that do not disrupt interaction flow [95].

### 5.3. Dual Use Risk and Intimate Partner Abuse

The most serious risk of the proposed agenda is not platform misuse but interpersonal misuse, specifically the potential for affective dating tools to be weaponized in contexts of intimate partner abuse and coercive control. This has been studied in various cases such as abusers exploiting consumer technology [24], as a mechanism of ongoign psychological control [37] among others.

This risk is compounded by the broader deployment of emotion AI in hgih-stakes contexts, including border control, employee performance monitoring, insurance assessment, where similar dual-use concerns have been extensively documented [17, 84]. What we propose is a set of architectural and design principles that reduce but cannot eliminate it: strict on-device processing and clear public documentation of misuse risks in any published sytem/dataset.

### 5.4. Regulatory Landscape and Open Legal Questions

Three open research questions guide responsible implementation. First, what consent mechanisms are technically and experientially appropriate for affective video processing in intimate communication? GDPR Article 9 classifies most emotion data as biometric special category information requiring explicit consent and Data Protection Impact Assessments, yet dating-specific UX remains underexplored [80]. Second, how can affective tools be architecturally designed to prevent platform-side misuse? Federated learning, on-device inference, and zero-knowledge proofs offer pathways to ensure emotional data never leaves the phone. Third, what regulatory gaps persist? While the EU AI Act [83] prohibits emotion recognition in workplaces and education, no jurisdiction explicitly governs its use in consensual dating contexts, leaving enforcement inconsistent globally [42].

### 5.5. The Limits of This Proposal

We close by being explicit about what this paper does not claim. We do not claim that affective computer vision will "solve" dating safety. It constitutes one tool within a broader sociotechnical ecosystem that must integrate platform policies (proactive moderation, easy blocking), legal frameworks (anti-harassment laws), and cultural norms (bystander intervention). We also do not claim that the technical capabilities claimed in section 3 are ready for deployment, they require new datasets and extensive fairness auditing. Current models also face technical limits: cultural bias, neurodiversity challenges, and imperfect accuracy, that no system can fully overcome without human oversight and ethical responsibility. The proposal is scoped specifically to early-stage, dyadic romantic interaction, not workplace, familial, or public settings.

## 6. Conclusion

Online dating has become the primary context in which romantic relationships begin, yet the platforms mediating these interactions remain perceptually impoverished, stripping the nonverbal cues that humans rely on to navigate attraction, discomfort, and consent. This paper has argued that this gap is neither inevitable nor neutral: it is a design failure with measurable consequences that fall disproportionately on women and vulnerable users. Addressing this issue lies squarely within the technical capabilities of the computer vision community. Mature affective CV tools, such as facial action unit detection, gaze estimation, engagement modeling, and multimodal affect recognition, already exist. What has been missing is a research agenda that treats the dating context seriously, positions fairness and consent as foundational requirements, and asks what these tools should be accomplishing rather than merely what they can do.

We have outlined such an agenda, grounded in the social psychology of dating and organized around discomfort detection, engagement asymmetry modeling, consent-aware interaction design, and longitudinal interaction summarization. Pursuing it responsibly demands new datasets to be collected under dyadic consent protocols, fairness evaluation disaggregated across race, gender identity, neurotype, and cultural background, and architectural commitments to on-device processing that prevent affective data from becoming platform surveillance infrastructure. The CV community has built much of the technical substrate that contemporary dating platforms already run on, largely without asking what responsibilities that entails.